\documentclass[11pt]{article}

\usepackage[preprint]{acl}

\usepackage{times}
\usepackage{latexsym}

\usepackage[T1]{fontenc}
\usepackage[utf8]{inputenc}

\usepackage{microtype}

\usepackage{inconsolata}

\usepackage{graphicx}
\usepackage{subfigure}
\usepackage{subcaption}
\usepackage{color}

\usepackage{mdframed}       % prompt showcase

\usepackage{amssymb}
\usepackage{amsmath}
\usepackage{multirow}%
\usepackage{amsmath,amssymb,amsfonts}%
\usepackage{amsthm}%
\usepackage{mathrsfs}%
\usepackage[title]{appendix}%
\usepackage{xcolor}%
\usepackage{cite}%
\usepackage{textcomp}%
\usepackage{manyfoot}%
\usepackage{booktabs}%
\usepackage{algorithm}%
\usepackage{algorithmicx}%
\usepackage{algpseudocode}%
\usepackage{listings}%
\usepackage{utfsym}%
\usepackage{tcolorbox}

\definecolor{mainboxbg}{HTML}{F7F9FC}
\definecolor{mainboxborder}{HTML}{A1C6EA}
\definecolor{adminboxbg}{HTML}{FFF4E5}
\definecolor{adminboxborder}{HTML}{F5A623}

\usepackage{indentfirst}
\usepackage[utf8]{inputenc}
\usepackage{makecell}

\title{CSC-SQL: Corrective Self-Consistency in Text-to-SQL \\
via Reinforcement Learning}

\author{Lei Sheng\footnotemark[1] \\\makecell{  Wuhan University\\ of Technology, China} \\  \texttt{xuanfeng1992@whut.edu.cn} \\
  \And
  Shuai-Shuai Xu \\ \makecell{ University of Science and \\Technology of China, China} \\ \texttt{sa517432@mail.ustc.edu.cn} \\
}
\usepackage{hyperref}

\begin{document}
\maketitle

\footnotetext[1]{Corresponding author: \texttt{xuanfeng1992@whut.edu.cn}}

\begin{abstract}
Large language models (LLMs) have demonstrated strong capabilities in translating natural language questions about relational databases into SQL queries. In particular, test-time scaling techniques such as Self-Consistency and Self-Correction can enhance SQL generation accuracy by increasing computational effort during inference. However, these methods have notable limitations: Self-Consistency may select suboptimal outputs despite majority votes, while Self-Correction typically addresses only syntactic errors. To leverage the strengths of both approaches, we propose CSC-SQL, a novel method that integrates Self-Consistency and Self-Correction. CSC-SQL selects the two most frequently occurring outputs from parallel sampling and feeds them into a merge revision model for correction. Additionally, we employ the Group Relative Policy Optimization (GRPO) algorithm to fine-tune both the SQL generation and revision models via reinforcement learning, significantly enhancing output quality. Experimental results confirm the effectiveness and generalizability of CSC-SQL. On the BIRD private test set, our 7B model achieves 71.72\% execution accuracy, while the 32B model achieves 73.67\%. The code has been open sourced at \url{https://github.com/CycloneBoy/csc_sql}.

\end{abstract}

\section{Introduction}
\label{sec:introduction}

The Text-to-SQL task involves translating natural language questions into SQL queries for database access \citep{katsogiannis-meimarakis_survey_2023, shi_survey_2024}. Current mainstream approaches rely on large language models (LLMs) to generate SQL \citep{liu_survey_2024}. Since LLMs typically employ stochastic sampling during inference, introducing additional computation at test time has been shown to improve output quality \citep{wang2023selfconsistencyimproveschainthought}. This family of techniques is collectively known as test-time scaling (TTS) \citep{zhang2025surveytesttimescalinglarge}. TTS methods have been widely adopted in domains such as mathematics, code generation, and logical reasoning. In Text-to-SQL task, the most prevalent TTS strategies are Self-Consistency (SC) \citep{dong_c3_2023,gao_text--sql_2023,xie2025opensearchsqlenhancingtexttosqldynamic} and Self-Correction \citep{volvovsky_dfin-sql_2024,cao2024rslsqlrobustschemalinking,xiyansql}. 

% grpo 7b scale 
\begin{figure}[t]
	\begin{center}
    \includegraphics[width=.5\textwidth]{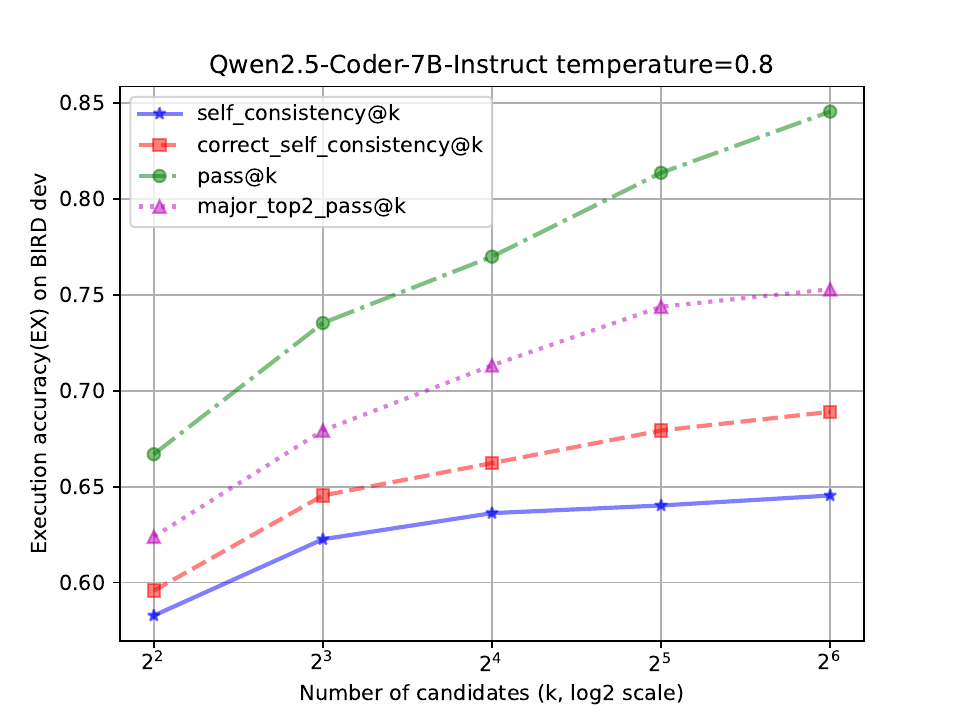}
		  \caption{On the BIRD development set, the trend chart of different metrics under different sampling numbers using the Qwen2.5-Coder-7B-Instruct model at a temperature of 0.8. Among them, major\_top2\_pass@k denotes selecting the two groups with the most votes to calculate the pass@k metric, and self\_consistency@k and correct\_self\_consistency@k represent the results of using the SC method and our CSC method respectively.}
		  \label{figure:7b_n_scale_t08} 
	\end{center}
\end{figure}

% overview 
\begin{figure*}[htbp]
	\begin{center}
		  \includegraphics[width=0.7\textwidth]{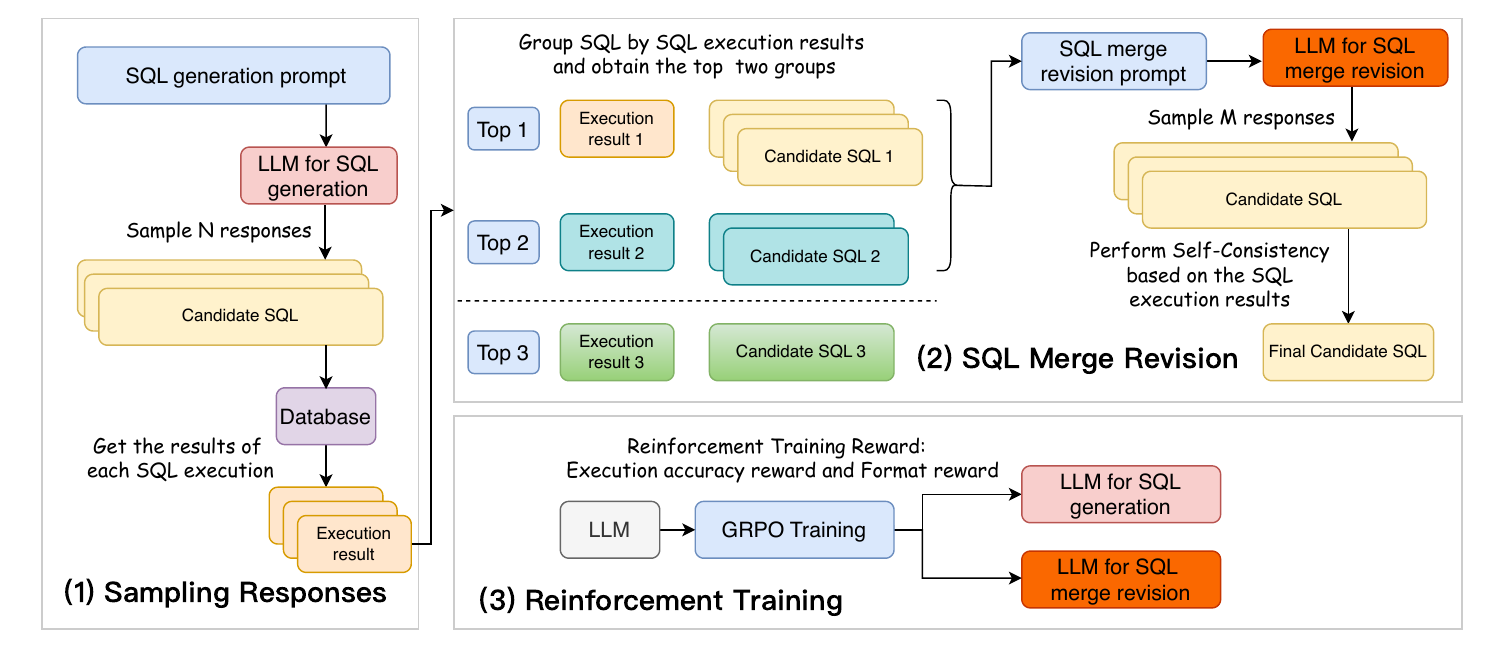}
		  \caption{Overview of the proposed CSC-SQL framework}
		  \label{figure:csc_sql} 
	\end{center}
\end{figure*}

A primary limitation of SC is that the most frequently selected output is not always the correct final result. As illustrated in Figure \ref{figure:7b_n_scale_t08}, the gap between self\_consistency@k and pass@k widens as the number of candidates increases. To address this, recent studies \citep{pourreza2024chasesqlmultipathreasoningpreference,xiyansql,gorti2024msc} have employed fine-tuned selection models to identify the final output. However, training effective selection models requires high-quality labeled data, which is currently scarce. To further investigate the behavior of SC, we propose a new metric, major\_top2\_pass@k, which calculates pass@k based only on the top two voting groups, determined by SQL execution results. As shown in Figure \ref{figure:7b_n_scale_t08}, major\_top2\_pass@k mitigates the gap between self\_consistency@k and pass@k by leveraging only the top two candidate groups.

To further enhance the SC method, we propose the Corrective Self-Consistency (CSC) framework CSC-SQL. First, the SQL generation model produces N candidate SQLs via parallel sampling. The execution results of these candidates are then used to identify the top two groups based on majority voting. These top two SQL queries, along with the merged schema and their execution outcomes, are used to construct a Merge-Revision template \citep{sheng2025basesqlpowerfulopensource}. This template is then passed to a revision model, which generates M new candidate SQL queries. Finally, the Self-Consistency method is applied again to select the final output. During the post-training of LLMs \citep{tie2025surveyposttraininglargelanguage}, reinforcement learning (RL) has been shown to significantly enhance reasoning capabilities \citep{besta2025reasoninglanguagemodelsblueprint}. Following the approach of \citep{pourreza2025reasoningsqlreinforcementlearningsql,ma2025sqlr1trainingnaturallanguage}, we post-train both the SQL generation model and the revision model using the Group Relative Policy Optimization (GRPO) \citep{shao2024deepseekmathpushinglimitsmathematical} algorithm, with execution accuracy (EX) and format consistency as the reward functions.

We conducted experiments on the BIRD \citep{li_can_2023} and Spider \citep{yu_spider_2019} datasets using models of varying sizes. The results on BIRD development set demonstrate that our proposed CSC method consistently improves execution accuracy by 0.72\%-5.54\% compared to the Self-Consistency baseline. Specifically, using the CSC-SQL framework, the 3B model achieved 65.28\% EX, the 7B model achieved 69.19\% EX, the 32B model achieved 71.33\% EX, and the 72B model achieved 71.69\% EX. On the BIRD private test set, our 7B model achieves 71.72\% EX, while the 32B model achieves 73.67\%. We tested the model trained on the BIRD dataset directly on the Spider dataset. The experimental results show that our CSC-SQL method has strong generalization ability and also brings certain performance improvements.

Our main contributions can be summarized as: (1) We analyze the limitations of the Self-Consistency method in the Text-to-SQL task and propose a novel framework, CSC-SQL, to address them. (2) We apply reinforcement learning techniques to enhance the SQL generation capability of LLMs, and train six models to verify the effectiveness of the approach. (3) CSC-SQL achieves competitive performance on the BIRD development and test set.

\section{Methodology}
\label{sec:methods}

Our CSC-SQL framework, illustrated in Figure \ref{figure:csc_sql}, consists of three main components: Sampling Responses, SQL Merge Revision and Reinforcement Training.

% correct self-consistency result
% \begin{center}
\begin{table*}[h]\tiny
  \centering
\begin{tabular*}{0.81\linewidth}{c|c|c|l|l|l|l|l}    
  \toprule

  \multicolumn{2}{c|}{\text{SQL Generation}}   & \multirow{2}*{\makecell{Post\\Method}}  &  \multicolumn{4}{c}{\text{Dev EX(\%)}} \\
  \cline{1-2} \cline{4-8}
  
  \text{Model}  & \text{Training} & & \text{n=4} & \text{n=8} & \text{n=16} & \text{n=32} & \text{n=64}\\ 
 \midrule

  % Qwen2.5-Coder-3B-Instruct
  \multirow{4}*{\text{\makecell{Qwen2.5-Coder-\\3B-Instruct}}}  & \multirow{2}*{\text{-}}  & \text{SC}  & \text{44.70} & \text{49.87} & \text{53.19} & \text{55.58} & \text{56.15} \\
  &   & \text{CSC}  & \text{45.96\tiny(+1.26)} & \text{51.43\tiny(+1.56)} & \text{55.91\tiny(+2.72)} & \text{58.84\tiny(+3.26)} & \text{60.39\tiny(+4.24)} \\
  \cline{2-8}
  & \multirow{2}*{\text{GRPO}}  & \text{SC}  & \text{53.72} & \text{56.95} & \text{59.06} & \text{59.93} & \text{60.60} \\
  &   & \text{CSC}  & \text{54.65\tiny(+0.93)} & \text{58.41\tiny(+1.46)} & \text{61.36\tiny(+2.30)} & \text{62.13\tiny(+2.20)} & \text{63.34\tiny(+2.74)} \\

  \midrule
  % XiYanSQL-QwenCoder-3B-2502
  \multirow{4}*{\text{\makecell{XiYanSQL-QwenCoder-\\3B-2502}}}  & \multirow{2}*{\text{-}}  & \text{SC}  & \text{49.22} & \text{54.56} & \text{56.82} & \text{58.19} & \text{58.87} \\
  &   & \text{CSC}  & \text{51.06\tiny(+1.85)} & \text{57.74\tiny(+3.17)} & \text{61.10\tiny(+4.28)} & \text{63.23\tiny(+5.04)} & \text{64.41\tiny(+5.54)} \\
  \cline{2-8}
  & \multirow{2}*{\text{GRPO}}  & \text{SC}   & \text{56.76} & \text{59.95} & \text{60.78} & \text{61.23} & \text{61.30} \\
  &   & \text{CSC}  & \text{58.15\tiny(+1.39)} & \text{62.17\tiny(+2.22)} & \text{63.49\tiny(+2.71)} & \text{64.91\tiny(+3.68)} & \text{65.28\tiny(+3.98)} \\
  
  \midrule
  % Qwen2.5-Coder-7B-Instruct
  \multirow{4}*{\text{\makecell{Qwen2.5-Coder-\\7B-Instruct}}}  & \multirow{2}*{\text{-}}  & \text{SC}  & \text{58.34} & \text{62.18} & \text{63.35} & \text{63.80} & \text{64.45} \\
  &   & \text{CSC}  & \text{59.63\tiny(+1.28)} & \text{64.28\tiny(+2.10)} & \text{66.41\tiny(+3.06)} & \text{67.54\tiny(+3.73)} & \text{68.70\tiny(+4.25)} \\
  \cline{2-8}
  & \multirow{2}*{\text{GRPO}}  & \text{SC}  & \text{63.73} & \text{64.95} & \text{65.58} & \text{65.88} & \text{66.15} \\
  &   & \text{CSC}  & \text{64.91\tiny(+1.17)} & \text{66.96\tiny(+2.01)} & \text{68.30\tiny(+2.72)} & \text{68.61\tiny(+2.73)} & \text{69.19\tiny(+3.04)} \\

  \midrule
  % XiYanSQL-QwenCoder-7B-2502
  \multirow{4}*{\text{\makecell{XiYanSQL-QwenCoder-\\7B-2502}}}  & \multirow{2}*{\text{-}}  & \text{SC} & \text{59.26} & \text{62.58} & \text{63.67} & \text{64.84} & \text{66.19} \\
  &   & \text{CSC}  & \text{61.13\tiny(+1.87)} & \text{65.34\tiny(+2.76)} & \text{66.93\tiny(+3.26)} & \text{68.08\tiny(+3.24)} & \text{69.21\tiny(+3.02)} \\
  \cline{2-8}
  & \multirow{2}*{\text{GRPO}}  & \text{SC}   & \text{61.41} & \text{63.71} & \text{64.65} & \text{65.69} & \text{65.95} \\
  &   & \text{CSC}  & \text{63.04\tiny(+1.63)} & \text{66.08\tiny(+2.37)} & \text{67.71\tiny(+3.06)} & \text{68.86\tiny(+3.17)} & \text{69.04\tiny(+3.09)} \\
  
  \midrule
  % OmniSQL-7B
  \multirow{2}*{\text{\makecell{OmniSQL-7B}}}  & \multirow{2}*{\text{-}}  & \text{SC} & \text{65.86}& \text{67.26} & \text{67.23} & \text{67.42}  & \text{67.54} \\
  &   & \text{CSC}  & \text{66.82\tiny(+0.96)} & \text{68.32\tiny(+1.06)} & \text{69.21\tiny(+1.99)} & \text{69.23\tiny(+1.81)}  & \text{69.62\tiny(+2.08)} \\
    
 \midrule
  % Meta-Llama-3.1-8B-Instruct
  \multirow{2}*{\text{\makecell{Meta-Llama-3.1-\\8B-Instruct}}}  & \multirow{2}*{\text{-}}  & \text{SC} & \text{47.74}& \text{52.35} & \text{55.74} & \text{56.89}  & \text{58.11} \\
  &   & \text{CSC}  & \text{49.15\tiny(+1.41)} & \text{54.65\tiny(+2.30)} & \text{58.80\tiny(+3.06)} & \text{61.45\tiny(+4.56)}  & \text{62.84\tiny(+4.74)} \\
  
  \midrule
  % gemma-3-12b-it
  \multirow{2}*{\text{\makecell{gemma-3-12b-it}}}  & \multirow{2}*{\text{-}}  & \text{SC} & \text{57.69}& \text{59.34} & \text{60.50} & \text{61.15}  & \text{60.89} \\
  &   & \text{CSC}  & \text{58.89\tiny(+1.20)} & \text{61.26\tiny(+1.91)} & \text{63.10\tiny(+2.61)} & \text{63.73\tiny(+2.59)}  & \text{63.95\tiny(+3.06)} \\
   
  \midrule
  % Qwen2.5-Coder-14B-Instruct
  \multirow{2}*{\text{\makecell{Qwen2.5-Coder-\\14B-Instruct}}}  & \multirow{2}*{\text{-}}  & \text{SC}  & \text{65.49} & \text{66.36} & \text{67.24} & \text{67.17} & \text{67.19} \\
  &   & \text{CSC}  & \text{66.51\tiny(+1.02)} & \text{68.02\tiny(+1.66)} & \text{69.28\tiny(+2.04)} & \text{69.49\tiny(+2.33)} & \text{69.60\tiny(+2.41)} \\

  \midrule
  % Qwen2.5-Coder-32B-Instruct
  \multirow{2}*{\text{\makecell{Qwen2.5-Coder-\\32B-Instruct}}}  & \multirow{2}*{\text{-}}  & \text{SC}  & \text{66.97} & \text{67.56} & \text{68.54} & \text{68.58} & \text{68.67} \\
  &   & \text{CSC}  & \text{68.02\tiny(+1.06)} & \text{69.12\tiny(+1.56)} & \text{69.95\tiny(+1.41)} & \text{70.32\tiny(+1.74)} & \text{70.69\tiny(+2.02)} \\

  \midrule
  % XiYanSQL-QwenCoder-32B-2412
  \multirow{2}*{\text{\makecell{XiYanSQL-QwenCoder-\\32B-2412}}}  & \multirow{2}*{\text{-}}  & \text{SC} & \text{66.11} & \text{67.50} & \text{68.45} & \text{68.75} & \text{68.95} \\
  &   & \text{CSC}  & \text{67.26\tiny(+1.15)} & \text{69.37\tiny(+1.88)} & \text{70.57\tiny(+2.12)} & \text{70.87\tiny(+2.13)} & \text{71.33\tiny(+2.39)} \\

  \midrule
  % Meta-Llama-3.1-70B
  \multirow{2}*{\text{\makecell{Meta-Llama-3.1-70B}}}  & \multirow{2}*{\text{-}}  & \text{SC} & \text{65.58} & \text{67.43} & \text{69.23} & \text{69.56} & \text{69.82} \\
  &   & \text{CSC}  & \text{66.30\tiny(+0.72)} & \text{68.86\tiny(+1.43)} & \text{70.75\tiny(+1.52)} & \text{71.45\tiny(+1.89)} & \text{71.69\tiny(+1.87)} \\

  \bottomrule
\end{tabular*}
\caption{The table shows the EX comparison results of different SQL generation models with different post-processing methods of SC and CSC under different sampling numbers on the BIRD development set. SC represents the direct use of Self-Consistency, and CSC represents the Corrective Self-Consistency method we proposed. The N represents the number of candidate results generated by the SQL generation model. The number of candidate results of the SQL merge revision model is fixed at 8. The 3B SQL generation model uses the 3B SQL merge revision model, and the models of other sizes all use the 7B SQL merge revision model.}
\label{tab:result_main_bird}
\footnotetext{}
\end{table*}
% \end{center}

\textbf{CSC-SQL framework:} First, the SQL generation model produces N candidate SQLs through parallel sampling. These candidates are then grouped and ranked based on their execution results, and the top two groups are selected. If the candidate SQL queries are consistent, they are directly used as the final output. Otherwise, the two selected SQL queries, along with their execution results, are used to construct a template following the method of \citep{sheng2025basesqlpowerfulopensource}. This template is input into the merge-revision model to generate M revised candidate SQL queries. Finally, the Self-Consistency method \citep{wang2023selfconsistencyimproveschainthought} is applied to these m candidates to select the final SQL output.

\textbf{Reinforcement Training:} We apply the GRPO \citep{shao2024deepseekmathpushinglimitsmathematical} algorithm to perform reinforcement training on both the SQL generation model and the merge-revision model, aiming to enhance their generation capabilities. The reward function comprises two components: execution accuracy reward and format reward \citep{pourreza2025reasoningsqlreinforcementlearningsql,ma2025sqlr1trainingnaturallanguage,papicchio2025think2sqlreinforcellmreasoning}. The execution accuracy reward  ($R_{EX}$) evaluates the correctness of the generated SQL queries by comparing their execution results with those of the gold-standard SQL on the target database. The format reward ($R_{Format}$) encourages the model to structure its output by including the reasoning process within <think>...</think> tags and the final answer within <answer>...</answer> tags.

\begin{align}
R_{EX}= 
\begin{cases} 
1, & \text{if execution results is correct.} \\
0, & \text{otherwise}
\end{cases}
\end{align}

\begin{align}
R_{Format}= 
\begin{cases} 
1, & \text{if output format is match.} \\
0, & \text{otherwise}
\end{cases}
\end{align}

The final reward is  a weighted sum of the execution accuracy reward and the format reward:
\begin{equation}
  R = R_{EX} + 0.1 * R_{Format}
\end{equation}

\textbf{SQL Merge Revision Dataset:} We utilized the Qwen2.5-Coder-7B-Instruct model and its GRPO-trained variant to generate eight candidate SQL queries in parallel on the BIRD training set. These candidates were grouped and voted upon based on their execution results. The two groups receiving the highest number of votes were selected to construct the merged correction template dataset. 
% If all eight candidate SQL queries were consistent, the data instance was discarded.

\section{Experiments}
\label{sec:experiments}

\subsection{Experiments Setting}
\label{subsec:experiments_setting}

We use the BIRD \citep{li_can_2023} and Spider \citep{yu_spider_2019} dataset as the evaluation datasets. Execution accuracy (EX) is adopted as the evaluation metric. Unless otherwise specified, all experiments use default sampling parameters, with a temperature of 0.8 and a merged revision sample size of 8. Results are reported as the average performance over three runs for each experimental setting. Additional implementation details are provided in Appendix \ref{sec:appendix_experiment_setting}.

\subsection{Main Results}
\label{subsec:main_result}

Our experimental results are presented in Table \ref{tab:result_main_bird}, they demonstrate that:

\textbf{For the same model and samples size n, the CSC method consistently outperforms SC method.}  Specifically, CSC achieves a steady improvement in EX ranging from 0.72\% to 5.54\% compared to SC. For instance, when the sample size is 64, the EX of XiYanSQL-QwenCoder-3B-2502 without GRPO post-training increases from 58.87\% to 64.41\%, while Meta-Llama-3.1-8B-Instruct improves from 58.11\% to 62.84\%. In most models, as the number of samples n increases, the performance gain of CSC over SC also increases. As illustrated in Figure \ref{figure:7b_n_scale_t08}, when the sample size increases from 4 to 64, the EX of Qwen2.5-Coder-7B-Instruct without GRPO training improves from 1.28\% to 4.25\%.

\textbf{For the same model, GRPO training generally improves EX under both SC and CSC methods. } For example, with a sample size of 8, Qwen2.5-Coder-3B-Instruct shows an EX improvement of 7.08\% under SC and 6.98\% under CSC, while Qwen2.5-Coder-7B-Instruct improves by 2.77\% under SC and 2.68\% under CSC. However, as the sample size n increases, the benefits of GRPO training diminish. At a sample size of 64, Qwen2.5-Coder-7B-Instruct achieves only a 1.70\% improvement under SC and 0.49\% under CSC.

\textbf{For a fixed sampling number n, models within the same family consistently achieve higher EX under the CSC method compared to the SC method, especially as model size increases.} For instance, when using the Qwen2.5-Coder model family with a sampling number of 32, EX improves from 58.84\% for the 3B model to 70.32\% for the 32B model. Moreover, CSC enables smaller language models to outperform larger counterparts within the same family. For example, with a sampling number of 64, Qwen2.5-Coder-7B-Instruct using CSC surpasses both Qwen2.5-Coder-14B-Instruct and Qwen2.5-Coder-32B-Instruct using SC. These results highlight the efficiency benefits of CSC, as it allows smaller models to substitute for larger ones while maintaining or improving performance.

\subsection{Comparison with other methods}
\label{sec:comparison_sota_methods}

Results on the BIRD dataset show that our CSC-SQL approach performs strongly competitively with previous Text-to-SQL approaches, as shown in Table \ref{tab:result_main_bird_other}. On the BIRD private test set, our 32B model (base model XiYanSQL-QwenCoder-32B-2412\citep{xiyansql} ) achieved an EX of 73.67\%, surpassing all other methods using open source models and improving by 4.64 points compared to the original base model XiYanSQL-QwenCoder-32B-2412; the 7B model (base model Qwen2.5-Coder-7B-Instruct) achieved an EX of 71.72\%, surpassing all other methods using the same base model, showing strong competitiveness. Detailed comparative analysis results are shown in Appendix \ref{sec:comparison_other_methods}.

% \textbf{Best results:} When the number of samples is 64, the CSC-SQL method is used. The 3B model achieves 65.28\% EX, the 7B model achieves 69.19\% EX, the 32B model achieves 71.33\% EX, and the 72B model achieves 71.69\% EX, indicating the effectiveness of the CSC-SQL framework.

\subsection{Analysis}
\label{subsec:analysis}

% diff version merge result
\begin{center}
\begin{table}[htbp]\tiny
  \setlength{\tabcolsep}{4pt}
  % \centering
\begin{tabular*}{0.6\linewidth}{c|c|c|c|c|c|c|c|c}    
  \cline{1-9}

  % Generation\\ Size (n) Revision\\ Size (m)
 \multicolumn{2}{c|}{\text{SQL Generation}}  & \multirow{3}*{\makecell{G\\ Size \\(n)}} & \multirow{3}*{\makecell{SC \\EX\\(\%)}} & \multirow{3}*{\makecell{R\\ Size \\(m)}} & \multicolumn{4}{c}{\text{CSC EX(\%)}} \\
  \cline{1-2} \cline{6-9}
    \multirow{2}*{\makecell{Model}} & \multirow{2}*{\makecell{Training}} & & & &  \multicolumn{2}{c|}{\makecell{Without GRPO}} &  \multicolumn{2}{c}{With GRPO} \\ 
  \cline{6-9}
  & & &  & &  \makecell{3B} &  \makecell{7B}  &  \makecell{3B} & \makecell{7B} \\ 
 \cline{1-9}

  % Qwen2.5-Coder-7B-Instruct
  \multirow{16}*{\text{\makecell{Qwen2.5-\\Coder-\\7B-\\Instruct}}}  & \multirow{8}*{\text{-}}  & \multirow{4}*{8}  & \multirow{4}*{61.86} 
           & 1 & \text{56.45} & \text{59.65} & \text{63.04} & \text{63.62} \\
  &   &   &  & 4 & \text{58.86} & \text{61.53} & \text{63.91} & \text{64.04} \\
  &   &   &  & 8 & \text{59.71} & \text{61.73} & \text{63.86} & \text{63.93}  \\
  &   &  &  & 16 & \text{60.23} & \text{62.45} & \text{63.84} & \text{64.04}  \\
  \cline{3-9}
  & & \multirow{4}*{16}  & \multirow{4}*{63.65} 
  & 1 & \text{53.91} & \text{58.86} & \text{64.66} & \text{65.64}  \\
  &   &   &  & 4 & \text{57.82} & \text{62.25} & \text{66.30} & \text{66.67}  \\
  &   &   &  & 8 & \text{59.77} & \text{62.25} & \text{66.47} & \text{66.78}  \\
  &   &  &  & 16 & \text{60.30} & \text{62.77} & \text{66.51} & \text{66.73}  \\
  \cline{2-9}
  % Qwen2.5-Coder-7B-Instruct GRPO
  &  \multirow{8}*{\text{GRPO}}  & \multirow{4}*{8}  & \multirow{4}*{65.02} & 1 & \text{58.47} & \text{62.90}  & \text{66.16} & \text{66.68}  \\
  &   &   &  & 4 & \text{62.84} & \text{63.95} & \text{66.84} & \text{67.01}   \\
  &   &   &  & 8 & \text{61.79}& \text{64.14}  & \text{66.88} & \text{66.91}  \\
  &   &  &  & 16 & \text{63.16} & \text{63.62} & \text{67.08} & \text{67.06}  \\
  \cline{3-9}
  & & \multirow{4}*{16}  & \multirow{4}*{65.17} & 1 & \text{58.01} & \text{63.10}& \text{66.94} & \text{67.60}  \\
  &   &   &  & 4 & \text{61.47} & \text{64.86} & \text{67.95} & \text{67.86}  \\
  &   &   &  & 8 & \text{62.90} & \text{64.79} & \text{67.88} & \text{67.93}  \\
  &   &  &  & 16 & \text{62.71} & \text{65.25} & \text{67.88} & \text{67.95}  \\

  \cline{1-9}
\end{tabular*}
\caption{The table shows the corresponding EX of the SQL merge revision model under different configuration parameters using the 3B and 7B Qwen2.5-Coder-Instruct models on the BIRD development set without GRPO training and after GRPO training. G size(n) and R size(m) represent the number of parallel samples of the SQL generation model and the merge revision model, respectively.}
\label{tab:result_diff_merge_n_m}
\footnotetext{}
\end{table}
\end{center}

To evaluate the impact of CSC across different merge revision model sizes and parallel sampling size, we conducted a detailed experimental, presented in Table \ref{tab:result_diff_merge_n_m}. A comparison between the merge revision model with and without GRPO training reveals that the model without GRPO often results in degraded performance, indicating its limited error correction capability. In contrast, GRPO training enhances the model's ability to correct errors. When the SQL generation model and the number of its samples remain fixed, and the merge revision model samples only once (i.e., no second-round SC voting is performed), the GRPO-trained merge revision model still demonstrates strong error correction capabilities.

Further performance gains can be achieved by increasing the number of samples in the merge revision model and applying a second SC voting step. However, once the number of merge revision samples increases to 4-16, the size of the merge revision model and the number of its parallel samples have minimal impact on the final execution accuracy (EX). In contrast, increasing the number of samples generated by the SQL generation model has a more substantial effect on EX. This suggests that the primary factor influencing CSC performance is the number of samples generated by the SQL generation model. A higher number of samples results in a larger value of major\_top2\_pass@k (as shown in Figure \ref{figure:7b_n_scale_t08}), thereby increasing the likelihood that the top two SQL candidates selected for merge revision are accurate. Consequently, this leads to improved performance of the SQL merge revision model. These findings also support the trend observed in Table \ref{figure:7b_n_scale_t08}: for most models, as the number of SQL generation samples n increases, the performance gains of CSC over SC also increase.

For more detailed analysis, please refer to Appendix \ref{sec:additional_analysis}.

\section{Conclusion}
\label{sec:conclusion}

In this paper, we address the limitations of the SC method in the Text-to-SQL task, specifically that the result with the most votes is not always the optimal one. We propose a novel RL-based framework, CSC-SQL, which improves upon SC by feeding the top two voted SQL candidates into a merge revision model for SQL regeneration. This framework is further enhanced through post-training using the GRPO algorithm, resulting in significant performance gains over the standard SC approach. We conduct extensive experiments on the BIRD and Spider datasets, demonstrating the effectiveness and generalizability of CSC-SQL. Notably, small models trained with CSC-SQL can outperform larger models of the same architecture using SC. In future work, we plan to integrate strategies such as Adaptive Self-Consistency and Soft Self-Consistency to improve the efficiency of parallel sampling.

\section{Limitations}
\label{sec:limitations_and_future_work}

The CSC-SQL method requires more computational resources than the standard Self-Consistency approach. After parallel sampling, it performs result merging and correction, which increases computational overhead. To mitigate this, methods such as Adaptive Self-Consistency \citep{aggarwal-etal-2023-lets} and Soft Self-Consistency \citep{wang2024softselfconsistencyimproveslanguage} can be incorporated to enhance sampling efficiency. This enhancement is already part of our future work plan. Additionally, CSC-SQL relies on sufficient diversity among sampled results; if the model generates mostly consistent outputs, the CSC-SQL method may not yield significant performance improvements.

\bibliography{custom}

\appendix

\newpage

% \section{Appendix}

\section{Related Work}
\label{sec:related_works}

Early Text-to-SQL method was mostly based on rules and fine-tuned small language models \citep{wang-etal-2020-rat,guo-etal-2019-towards}. With the revolutionary progress of LLMs, the mainstream method now uses the powerful in-context learning capabilities of LLM to generate SQL \citep{dong_c3_2023,pourreza_din-sql_2023,gao_text--sql_2023,lee_mcs-sql_2024}. The pipeline method is adopted to integrate multiple sub-steps such as schema linking, SQL generation, and Self-Correction to improve the accuracy of generation \citep{talaei_chess_2024,pourreza2024chasesqlmultipathreasoningpreference,xiyansql,xie2025opensearchsqlenhancingtexttosqldynamic}. \citep{pourreza-rafiei-2024-dts,li_codes_2024,gorti2024msc,sheng2025basesqlpowerfulopensource} considering issues such as data privacy and execution efficiency, the method of fine-tuning open source models is adopted. With the rise of test time compute \citep{zhang2025surveytesttimescalinglarge} methods, it can significantly improve LLMs performance by adding additional computation during inference. \citep{gao_text--sql_2023, li_dawn_2024,xie2025opensearchsqlenhancingtexttosqldynamic} directly use Self-Consistency to select the final SQL from many candidate SQLs, while \citep{pourreza2024chasesqlmultipathreasoningpreference,xiyansql} train a selection model to chose the final SQL, thereby improving the consistency of model generation. \citep{yuan2025mctssqleffectiveframeworktexttosql, lyu2025sqlo1selfrewardheuristicdynamic, li2025alphasqlzeroshottexttosqlusing} use Monte Carlo Tree Search (MCTS) to dynamically expand the reasoning steps to generate SQL. \citep{he2025starsqlselftaughtreasonertexttosql,li2025omnisqlsynthesizinghighqualitytexttosql} uses high-quality rationales to perform supervised fine-tune (SFT) on the model, so that the model can first generate an effective reasoning process and then generate the final SQL. Recent reasoning language models \citep{besta2025reasoninglanguagemodelsblueprint,openai2024openaio1card,deepseekai2025deepseekr1incentivizingreasoningcapability} use RL-based post-training methods to promote the reasoning and reflection process inside the model, so that the model shows strong performance in complex reasoning tasks. \citep{pourreza2025reasoningsqlreinforcementlearningsql,ma2025sqlr1trainingnaturallanguage,papicchio2025think2sqlreinforcellmreasoning} uses the GRPO \citep{shao2024deepseekmathpushinglimitsmathematical} algorithm to post-train the model to improve the model's ability to generate SQL.

% main result
\begin{table}[htbp]\tiny
  \setlength{\tabcolsep}{4pt}
  \centering
\begin{tabular}{c|c|c|c|c}    
  \toprule
  \text{Method} & \text{Model} & \text{Size} & \makecell[c]{Dev \\EX(\%)} & \makecell[c]{Test \\EX(\%)} \\
  \midrule % unk
  \makecell[l]{CHASE-SQL \\ \citep{pourreza2024chasesqlmultipathreasoningpreference}} & \text{Gemini-1.5-pro} & \text{UNK} &\text{73.01} & \text{73.0}\\
  \makecell[l]{XiYan-SQL\\ \citep{xiyansql}} & \text{GPT-4o} & \text{UNK} &\text{73.34} & \text{75.63}\\
  \makecell[l]{RSL-SQL\\ \citep{cao2024rslsqlrobustschemalinking}} & \text{GPT-4o} & \text{UNK} &\text{67.21} &\text{68.70}\\

  \midrule %32b
  \makecell[l]{Reasoning-SQL\\ \citep{pourreza2025reasoningsqlreinforcementlearningsql}} & \text{Qwen2.5-Coder-14B-Instruct} & \text{14B} &\text{72.29} &\text{72.78}\\

  \makecell[l]{Alpha-SQL\\ \citep{li2025alphasqlzeroshottexttosqlusing}} & \text{Qwen2.5-Coder-32B-Instruct} & \text{32B} &\text{69.70} &\text{70.26}\\
  \makecell[l]{BASE-SQL\\ \citep{sheng2025basesqlpowerfulopensource}} & \text{Qwen2.5-Coder-32B-Instruct} & \text{32B} &\text{67.47} &\text{-}\\

  \makecell[l]{XiYan-SQL\\ \citep{xiyansql}} & \makecell[c]{XiYanSQL-QwenCoder\\-32B-2412} & \text{32B} &\text{67.01} &\text{69.03}\\

  \midrule %7b
  \makecell[l]{Reward-SQL \\ \citep{zhang2025rewardsqlboostingtexttosqlstepwise}} & \text{Qwen2.5-Coder-7B-Instruct} & \text{7B} &\text{68.90} &\text{-}\\

  \makecell[l]{Reasoning-SQL\\ \citep
  {pourreza2025reasoningsqlreinforcementlearningsql}} & \text{Qwen2.5-Coder-7B-Instruct} & \text{7B} &\text{68.05} &\text{-}\\

  \makecell[l]{Alpha-SQL\\ \citep{li2025alphasqlzeroshottexttosqlusing}} & \text{Qwen2.5-Coder-7B-Instruct} & \text{7B} &\text{66.80} &\text{-}\\

  \makecell[l]{SQL-R1 \\ \citep{ma2025sqlr1trainingnaturallanguage}} & \text{Qwen2.5-Coder-7B-Instruct} & \text{7B} &\text{66.60} &\text{-}\\

  \makecell[l]{OMNI-SQL \\ \citep{li2025omnisqlsynthesizinghighqualitytexttosql}} & \text{Qwen2.5-Coder-7B-Instruct} & \text{7B} &\text{66.10} &\text{67.97}\\
  
  \midrule % our
  \makecell[l]{CSC-SQL} & \text{Meta-Llama-3.1-70B} & \text{70B} &\text{71.69} &\text{-}  \\
  \makecell[l]{CSC-SQL} & \makecell[c]{XiYanSQL-QwenCoder\\-32B-2412} & \text{32B} &\text{71.33} &\text{73.67} \\
  \makecell[l]{CSC-SQL} & \text{Qwen2.5-Coder-32B-Instruct} & \text{32B} &\text{70.69} &\text{-} \\
  \makecell[l]{CSC-SQL} & \text{Qwen2.5-Coder-14B-Instruct} & \text{14B} &\text{69.60} &\text{-} \\
  \makecell[l]{CSC-SQL} & \text{OmniSQL-7B} & \text{7B} &\text{69.62} &\text{-} \\
  \makecell[l]{CSC-SQL} & \makecell[c]{XiYanSQL-QwenCoder\\-7B-2502} & \text{7B} &\text{69.21} &\text{-} \\
  \makecell[l]{CSC-SQL} & \text{Qwen2.5-Coder-7B-Instruct} & \text{7B} &\text{69.19} &\text{71.72} \\
  \makecell[l]{CSC-SQL} & \makecell[c]{XiYanSQL-QwenCoder\\-3B-2502} & \text{3B} &\text{65.28} &\text{-} \\
  \makecell[l]{CSC-SQL} & \text{Qwen2.5-Coder-3B-Instruct} & \text{3B} &\text{63.34} &\text{-}  \\

  \bottomrule
\end{tabular}
\caption{Performance Comparison of different Text-to-SQL methods on BIRD dev  and private test dataset.}
\label{tab:result_main_bird_other}
\footnotetext{}
\end{table}

% spider correct self-consistency result
% \begin{center}
\begin{table*}[htbp]\tiny
  \centering
\begin{tabular*}{0.875\linewidth}{c|c|c|c|l|l|l|l|l}    
  \toprule

  \multirow{2}*{\text{Dataset}}  & \multirow{2}*{\text{SQL Generateion Model}}  & \multirow{2}*{\makecell{Train \\Method}}   & \multirow{2}*{\makecell{Post\\Method}}  &  \multicolumn{4}{c}{\text{EX(\%)}} \\
  \cline{5-9}
  & & & & \text{n=4} & \text{n=8} & \text{n=16} & \text{n=32} & \text{n=64}\\ 
 \midrule

  % spider dev 
  % Qwen2.5-Coder-3B-Instruct
  \multirow{8}*{\text{\makecell{Spider Dev}}} & \multirow{4}*{\text{\makecell{Qwen2.5-Coder-\\3B-Instruct}}}  & \multirow{2}*{\text{-}}  & \text{SC}  & \text{64.41} & \text{68.09} & \text{70.21} & \text{70.70} & \text{70.79} \\
  & &   & \text{CSC}  & \text{64.60\tiny(+0.19)} & \text{68.76\tiny(+0.67)} & \text{70.70\tiny(+0.49)} & \text{71.28\tiny(+0.58)} & \text{72.15\tiny(+1.36)} \\
  \cline{3-9}
  & & \multirow{2}*{\text{GRPO}}  & \text{SC}  & \text{71.18} & \text{73.31} & \text{73.11} & \text{75.44} & \text{76.02} \\
  & &   & \text{CSC}  & \text{71.37\tiny(+0.19)} & \text{73.98\tiny(+0.67)} & \text{73.50\tiny(+0.39)} & \text{75.92\tiny(+0.48)} & \text{76.69\tiny(+0.67)} \\

  \cline{2-9}
  % XiYanSQL-QwenCoder-3B-2502
  & \multirow{4}*{\text{\makecell{XiYanSQL-QwenCoder-\\3B-2502}}}  & \multirow{2}*{\text{-}}  & \text{SC}  & \text{69.14} & \text{72.92} & \text{75.44} & \text{76.98} & \text{77.85} \\
  & &   & \text{CSC}  & \text{71.37\tiny(+2.23)} & \text{74.18\tiny(+1.26)} & \text{76.79\tiny(+1.35)} & \text{77.95\tiny(+0.97)} & \text{78.72\tiny(+0.87)} \\
  \cline{3-9}
  & & \multirow{2}*{\text{GRPO}}  & \text{SC}   & \text{74.37} & \text{76.01} & \text{75.24} & \text{75.92} & \text{76.21} \\
  & &   & \text{CSC}  & \text{74.47\tiny(+0.1)} & \text{76.21\tiny(+0.2)} & \text{75.44\tiny(+0.2)} & \text{76.40 \tiny(+0.48)} & \text{76.11\tiny(-0.1)} \\
  \midrule
  % spider test
% Qwen2.5-Coder-3B-Instruct
  \multirow{8}*{\text{\makecell{Spider Test}}} & \multirow{4}*{\text{\makecell{Qwen2.5-Coder-\\3B-Instruct}}}  & \multirow{2}*{\text{-}}  & \text{SC}  & \text{69.12} & \text{72.57} & \text{73.82} & \text{74.80} & \text{75.59} \\
  & &   & \text{CSC}  & \text{69.77\tiny(+0.65)} & \text{73.50\tiny(+0.93)} & \text{75.55\tiny(+1.73)} & \text{76.39\tiny(+1.59)} & \text{76.99\tiny(+1.40)} \\
  \cline{3-9}
  & & \multirow{2}*{\text{GRPO}}  & \text{SC}  & \text{74.20} & \text{75.78} & \text{76.99} & \text{76.48} & \text{77.50} \\
  & &   & \text{CSC}  & \text{74.57\tiny(+0.37)} & \text{76.53\tiny(+0.75)} & \text{77.60\tiny(+0.61)} & \text{77.50\tiny(+1.02)} & \text{78.16\tiny(+0.66)} \\

  \cline{2-9}
  % XiYanSQL-QwenCoder-3B-2502
  & \multirow{4}*{\text{\makecell{XiYanSQL-QwenCoder-\\3B-2502}}}  & \multirow{2}*{\text{-}}  & \text{SC}  & \text{72.29} & \text{76.15} & \text{76.67} & \text{77.27} & \text{78.06} \\
  & &   & \text{CSC}  & \text{73.22\tiny(+0.93)} & \text{76.94\tiny(+0.79)} & \text{78.06\tiny(+1.39)} & \text{78.76\tiny(+1.49)} & \text{79.37\tiny(+1.31)} \\
  \cline{3-9}
  & & \multirow{2}*{\text{GRPO}}  & \text{SC}   & \text{75.45} & \text{76.43} & \text{76.99} & \text{77.04} & \text{77.69} \\
  & &   & \text{CSC}  & \text{76.06\tiny(+0.61)} & \text{77.08\tiny(+0.65)} & \text{77.92\tiny(+0.93)} & \text{78.20\tiny(+1.16)} & \text{78.57\tiny(+0.88)} \\

\bottomrule
\end{tabular*}
\caption{The table shows the EX comparison results of different SQL generation models with different post-processing methods of SC and CSC under different sampling numbers on the Spider development and test set. SC represents the direct use of Self-Consistency, and CSC represents the Corrective Self-Consistency method we proposed. The N represents the number of candidate results generated by the SQL generation model. The number of candidate results of the SQL merge revision model is fixed at 8. The 3B SQL generation model uses the 3B SQL merge revision model.}
\label{tab:result_main_spider}
\footnotetext{}
\end{table*}
% \end{center}

\section{Experiments Setting}
\label{sec:appendix_experiment_setting}

To evaluate the effectiveness of our CSC-SQL framework and the CSC method, we conducted experiments on 11 open-source LLMs. These models encompass diverse architectures and training strategies, enabling a comprehensive assessment of the adaptability of our approach. We employed the GRPO algorithm to fine-tune four different SQL generation models and two merge revision models of varying sizes (Qwen2.5-Coder-3B-Instruct and Qwen2.5-Coder-7B-Instruct, respectively).

We use the TRL \citep{vonwerra2022trl} library for GRPO training, employing a cosine learning rate scheduler with a warmup ratio of 0.1 and a small learning rate of 3e-6. Each model is trained for one epoch with an effective batch size of 12. For each input prompt, we generate six completions to satisfy the group size requirement for GRPO training. The maximum input template length is set to 8192 tokens, and the maximum generated output length is 1024 tokens. The execution accuracy reward function is assigned a weight of 1.0, while the format reward function is weighted at 0.1. All experiments are conducted on a machine equipped with four NVIDIA GPUs (each with 80 GB VRAM), and the vLLM \citep{kwon2023efficientmemorymanagementlarge} framework is used to accelerate inference during testing.

The method used to represent database table schemas follows a similar approach to previous work \citep{li_codes_2024,talaei_chess_2024}. Specifically, we adopt the representation based on CREATE TABLE statements as proposed by OmniSQL \citep{li2025omnisqlsynthesizinghighqualitytexttosql}, which incorporates column descriptions, representative values, and question-relevant values in the comment field.

% diff merge revision top k group size
\begin{center}
\begin{table}[htbp]\tiny
  \setlength{\tabcolsep}{5pt}
  % \centering
\begin{tabular*}{0.6\linewidth}{c|c|c|c|c|c|c|c}    
  \cline{1-8}

  % Generation\\ Size (n) Revision\\ Size (m)
  \multirow{3}*{\makecell{SQL \\Generation\\ Model}} & \multirow{3}*{\makecell{Train\\Method}}   & \multirow{3}*{\makecell{G\\ Size \\(n)}} & \multirow{3}*{\makecell{SC \\EX\\(\%)}} & \multirow{3}*{\makecell{R\\ Size \\(m)}} & \multirow{3}*{\makecell{Merge\\top k group \\ size }} & \multicolumn{2}{c}{\text{CSC EX(\%)}} \\
  \cline{7-8}
    & & &  & & &  \multicolumn{2}{c}{With GRPO} \\ 
  \cline{7-8}
  & & &  & & & \makecell{3B} & \makecell{7B} \\ 
 \cline{1-8}

  % Qwen2.5-Coder-7B-Instruct
  \multirow{12}*{\text{\makecell{Qwen2.5-\\Coder-\\7B-\\Instruct}}}  & \multirow{6}*{\text{-}}  & \multirow{3}*{8}  & \multirow{3}*{61.86} & \multirow{3}*{8} & 1 &  \text{62.32} & \text{62.58}  \\
  &   &  &  & & 2  & \textbf{63.86} & \textbf{63.93} \\
  &   &  & &  & 3  & \text{63.16} & \text{63.03}  \\
  \cline{3-8}
  & & \multirow{3}*{16}  & \multirow{3}*{63.65} & \multirow{3}*{8}  & 1 & \text{62.77} & \text{62.84}  \\
  &   &   &  & & 2  & \textbf{66.47} & \textbf{66.78}  \\
  &   &   &  & & 3  & \text{63.42} & \text{63.95}  \\
  \cline{2-8}
  % Qwen2.5-Coder-7B-Instruct GRPO
  &  \multirow{6}*{\text{GRPO}}  & \multirow{3}*{8}  & \multirow{3}*{65.02} &  \multirow{3}*{8} & 1  & \text{64.73} & \text{64.73}  \\
  &   &   &   & & 2  & \textbf{66.88} & \textbf{66.91}   \\
  &   &   &  &  & 3  & \text{65.12} & \text{65.05}  \\
  \cline{3-8}
  & & \multirow{3}*{16}  & \multirow{3}*{65.17} &  \multirow{3}*{8} & 1 & \text{65.44} & \text{65.38}  \\
  &   &   &  & & 2  & \textbf{67.88} & \textbf{67.93}   \\
  &   &   &  &  & 3 & \text{65.90} & \text{65.97}  \\

  \cline{1-8}
\end{tabular*}
\caption{The table shows the corresponding EX of the SQL merge revision model under different merge revision top K group sizes using the 3B and 7B Qwen2.5-Coder-Instruct models on the BIRD development set. G size(n) and R size(m) represent the number of parallel samples of the SQL generation model and the merge revision model, respectively.}
\label{tab:result_diff_merge_revision_topk_group_size}
\footnotetext{}
\end{table}
\end{center}

\section{Comparison with other methods}
\label{sec:comparison_other_methods}

Table \ref{tab:result_main_bird_other} shows the comparison results of the CSC-SQL method and other Text-to-SQL methods on the BIRD dataset. Our 7B model surpasses all other methods that utilize the same Qwen2.5-Coder-7B-Instruct model for post-training, achieving 69.19\% EX on the BIRD development set. Notably, Reward-SQL \citep{zhang2025rewardsqlboostingtexttosqlstepwise}, Reasoning-SQL \citep{pourreza2025reasoningsqlreinforcementlearningsql}, and SQL-R1 \citep{ma2025sqlr1trainingnaturallanguage} also employ the GRPO algorithm for post-training, further validating the effectiveness of CSC-SQL. Our 32B model achieves 71.33\% EX on the BIRD development set, outperforming other methods that use 32B-sized models \citep{li2025alphasqlzeroshottexttosqlusing,sheng2025basesqlpowerfulopensource,xiyansql} and trailing Reasoning-SQL \citep{pourreza2025reasoningsqlreinforcementlearningsql} by only 0.96\%. Reasoning-SQL incorporates both the GRPO algorithm and the CHASE-SQL \citep{pourreza2024chasesqlmultipathreasoningpreference} integration framework, and represents the best known method based on open-source models. Our 72B model achieves 71.69\% EX on the BIRD development set, only 2.77\% lower than the current state-of-the-art method, CHASE-SQL, which uses the proprietary LLM Gemini-1.5-Pro. This highlights the performance gap that still exists between open-source and closed-source LLMs. Importantly, our CSC-SQL method can also be integrated into other approaches as a replacement for the Self-Consistency component. Exploring such integrations will be a focus of our future work.

\section{Additional Analysis}
\label{sec:additional_analysis}

% bird sep result
% \begin{center}
\begin{table*}[htbp]\small
  \centering
\begin{tabular*}{0.7\linewidth}{c|c|c|l|l|l|l}    
  \toprule

  \multirow{2}*{\text{Model}}  & \multirow{2}*{\makecell{Train\\Method}}   & \multirow{2}*{\makecell{Post\\Method}}  &  \multicolumn{4}{c}{\text{Dev EX(\%)}} \\
  \cline{4-7}
  & & & \text{Simple} & \text{Moderate} & \text{Challenge} & \text{All} \\ 
  \midrule

  % Qwen2.5-Coder-3B-Instruct
  \multirow{4}*{\text{\makecell{Qwen2.5-Coder-\\3B-Instruct}}}  & \multirow{2}*{\text{-}}  & \text{SC}  & \text{64.11} & \text{45.91} & \text{38.62} & \text{56.19}  \\
  &   & \text{CSC}  & \text{67.89} & \text{50.43 } & \text{42.76} & \text{60.23}  \\
  \cline{2-7}
  & \multirow{2}*{\text{GRPO}}  & \text{SC}  & \text{68.65} & \text{49.14} & \text{45.52} & \text{60.56}  \\
  &   & \text{CSC} & \text{71.46} & \text{52.80} & \text{46.90} & \text{63.49}  \\
  \midrule
  % Qwen2.5-Coder-7B-Instruct
  \multirow{4}*{\text{\makecell{Qwen2.5-Coder-\\7B-Instruct}}}  & \multirow{2}*{\text{-}}  & \text{SC}  & \text{71.14 } & \text{56.47} & \text{48.28 } & \text{64.54}  \\
  &   & \text{CSC}  & \text{74.38} & \text{63.36 } & \text{51.72} & \text{68.90}  \\
  \cline{2-7}
  & \multirow{2}*{\text{GRPO}}  & \text{SC}  & \text{71.03} & \text{61.21 } & \text{52.41 } & \text{66.30}  \\
  &   & \text{CSC} & \text{73.95} & \text{64.01} & \text{55.86} & \text{69.23}  \\

  \bottomrule
\end{tabular*}
\caption{The table shows the performance of different models at different difficulty levels on the BIRD development set when the sampling temperature is 0.8 and the SQL generation model samples 64 candidate results in parallel.}
\label{tab:result_main_sep_bird}
\footnotetext{}
\end{table*}
% \end{center}

\subsection{Performance on Spider Dataset}
\label{sec:spider_main}

We evaluate the generality of the proposed CSC-SQL by evaluating it on the Spider dataset, directly using the model trained on the BIRD dataset without retraining a new model. Table \ref{tab:result_main_spider} presents the experimental results of the 3B model, demonstrating that the CSC method consistently outperforms the SC method. This highlights the strong generalization ability of the merge revision model.

\subsection{Analysis at different merge revision top k group sizes}
\label{sec:analysis_with_diff_topk}

To analyze the impact of different top k group sizes on merge revision model, we evaluated 7B model on the BIRD development set. We fixed the number of merge revision candidates to 8 and tested group sizes of 1, 2, and 3. The experimental results are presented in Table \ref{tab:result_diff_merge_revision_topk_group_size}. Among the configurations, a group size of 2 yields the best performance and is used as our default setting. A group size of 1 corresponds to standard revision mode, where each example is corrected individually; however, this results in limited improvement and, in some cases, performance degradation. When the group size is increased to 3, performance improves slightly but not significantly.

\subsection{Difficulty Analysis}
\label{sec:difficulty_analysis}

Table \ref{tab:result_main_sep_bird} presents the performance of the SC and our proposed CSC methods across different difficulty levels on the BIRD development set. The results indicate that the CSC method consistently outperforms the SC method at all difficulty levels.

\subsection{Detailed Cost Analysis}
\label{sec:analysis_cost}

% cost and performance result
\begin{table*}[htbp]\tiny
  \setlength{\tabcolsep}{4pt}
  \centering
\begin{tabular}{c|c|c|c|c|c|c|c|c|c|c}    
  \toprule
  \text{Model} & \text{Model Size} & \makecell{Train\\Method} & \text{SC Size} & \makecell{SC Use \\time (Hour)} & \makecell{CSC Use \\Time (Hour)} & \makecell{The proportion \\ of time that \\CSC increases (\%)} & \makecell{Total Use \\ Time (Hour)} & \makecell{Total \\Cost (\$) } & \makecell{Average Cost\\ Per Question (\$)} & \makecell[c]{Bird \\Dev EX (\%)}  \\
  \toprule
  % 7B
  \multirow{5}*{\text{\makecell{Qwen2.5-\\Coder-\\7B-\\Instruct}}} & \multirow{5}*{\text{7B}} & \multirow{5}*{\text{GRPO}}
      & 4 & 0.1 & 0.06 & 58 & 0.16 & 0.32 & 0.00022 & 64.91 \\
  & & & 8 & 0.14 & 0.07 & 52 &  0.21 & 0.42 & 0.00028 & 66.96 \\
  & & & 16 & 0.25 & 0.08 & 33 & 0.33 & 0.66 & 0.00044 & 68.30 \\
  & & & 32 & 0.47 & 0.08 & 17 & 0.55 & 1.1 & 0.00072 & 68.61 \\
  & & & 64 & 0.97 & 0.11 & 11 & 1.08 & 2.16 & 0.0014 & 69.19 \\
  \midrule % 
  % 32B
  \multirow{5}*{\text{\makecell{XiYanSQL-\\QwenCoder-\\32B-2412}}} & \multirow{5}*{\text{32B}} & \multirow{5}*{\text{-}} 
      & 4  & 0.2  & 0.07 & 33 & 0.27 & 0.54 & 0.00035 & 67.26 \\
  & & & 8  & 0.63 & 0.07 & 11 & 0.7 & 1.4 & 0.00091 & 69.37 \\
  & & & 16 & 1.0  & 0.07 & 7 & 1.07 & 2.14 & 0.0014 & 70.57 \\
  & & & 32 & 1.74 & 0.1 & 6 & 1.84 & 3.68 & 0.0024 & 70.87 \\
  & & & 64 & 3.18 & 0.1 & 3 & 3.28 & 6.56 & 0.0043 & 71.33 \\

  \bottomrule
\end{tabular}
\caption{The table shows the cost and performance of SQL generation on the BIRD development set using our CSC-SQL method with 7B and 32B sizes. The number of candidate SQLs generated by the merge revision model is fixed to 8.
The inference cost analysis uses an NVIDIA H800 with 80GB of memory (rented at \$2.0 per hour) as an example.}
\label{tab:result_cost_and_performance}
\footnotetext{}
\end{table*}

When using LLM-based algorithms, computational time and cost are critical considerations. The proposed CSC-SQL method is built on open-source models that require GPU servers for deployment in real-world applications. As an example of inference cost, we consider an H800 GPU server equipped with a single NVIDIA 80GB GPU (hardware configuration: 20-core Xeon  Platinum 8458P CPU, 100GB RAM, 500GB SSD, with a rental price of approximately \$2.0 per hour). We evaluated the inference time and cost of 7B and 32B models on the BIRD development set (containing 1,534 questions), and the results are presented in Table \ref{tab:result_cost_and_performance}. Since the merge revision model is 7B and generates 8 correction candidates, the total inference time increases gradually with the number of candidates generated by the SC method, while the proportion of time spent on the merge revision step decreases. For example, when the SQL generation model is 32B and generates 16 candidates, the additional inference time from the merge revision model increases by only about 7\%, resulting in an average cost of \$0.0014 per question. This demonstrates an effective balance between performance and cost. For performance-focused applications, the number of SQL generation candidates can be increased to 64. Overall, CSC-SQL offers flexible trade-offs between computational efficiency and model performance.

\subsection{Analysis at different sampling temperatures}
\label{sec:analysis_with_diff_temperatures}

\begin{figure*}
	\centering
	\begin{minipage}[b]{0.95\textwidth}
		\subfigure{
			\includegraphics[width=0.5\textwidth]{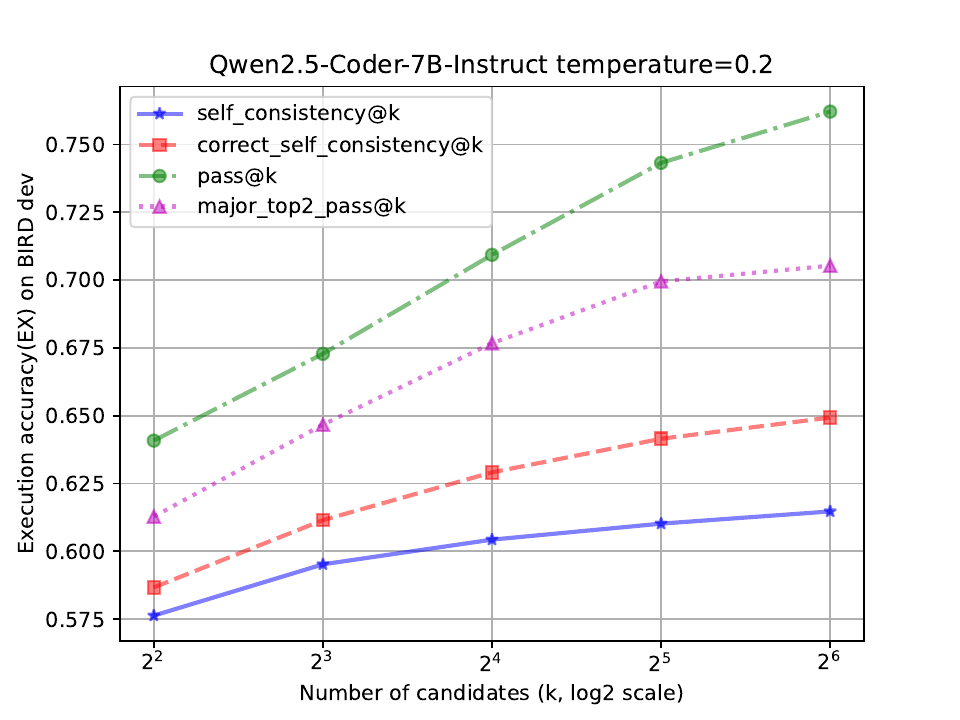} 
      
			\includegraphics[width=0.5\textwidth]{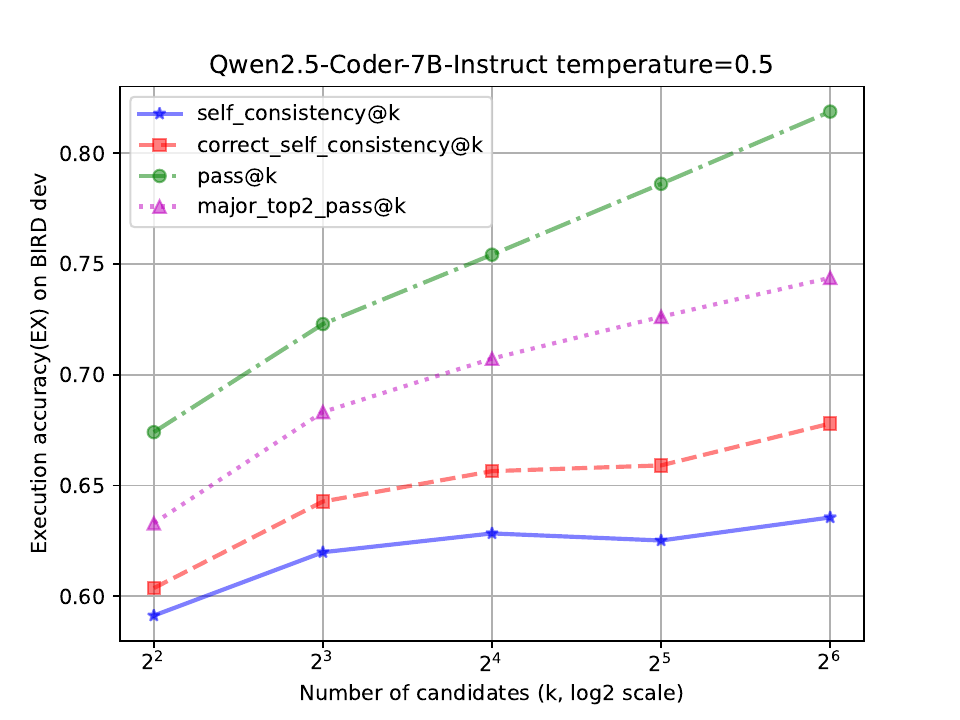}
			\label{fig:bash_calibration}
	}
	\end{minipage}
	\begin{minipage}[b]{0.95\textwidth}
		\subfigure{
			\includegraphics[width=0.5\textwidth]{Qwen2.5-Coder-7B-Instruct_temperature_0.8.pdf} 
			\includegraphics[width=0.5\textwidth]{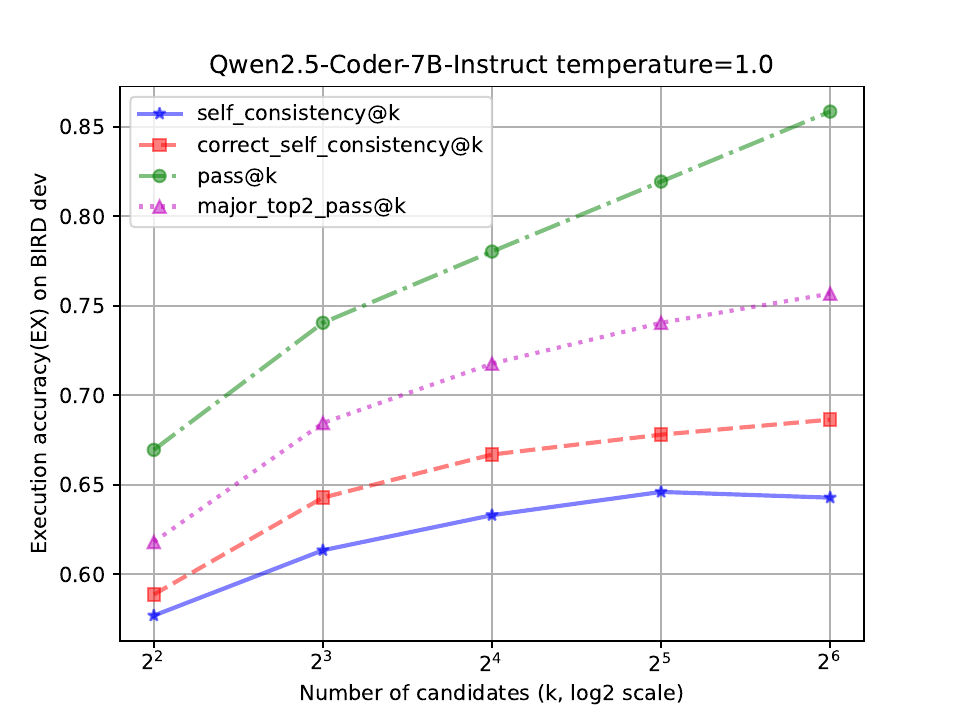}
			\label{fig:webshop_calibration}
	}
	\end{minipage}
	\caption{On the BIRD development set, using the Qwen2.5-Coder-7B-Instruct model, when the sampling temperature increases from 0.2 to 1.0, the trend charts of different metric under different sampling numbers.}
	\label{fig:7b_n_scale_t}
 \vspace{-0.5em}
\end{figure*}

The self-consistency method determines the final result by voting among multiple sampled candidates. However, when the diversity of these candidates is limited, the effectiveness of voting diminishes. Similar to the self-consistency approach, our proposed CSC method also relies on the diversity of sampled results. To investigate this, we tested four sampling temperatures ranging from 0.2 to 1.0. As shown in Figure \ref{fig:7b_n_scale_t}, regardless of the sampling temperature, all evaluation metrics improve as the number of samples increases. At lower temperatures, limited diversity leads to lower metric scores. When the temperature is set to 0.8 or 1.0, the performance difference is minimal. Therefore, we adopt a temperature of 0.8 in subsequent experiments.

% figure scale temperature  [htbp]

\section{Output Example Analysis}
\label{sec:output_example_analysis}

In this section, we present example output results from the SQL generation model using the Qwen2.5-Coder-7B model trained with GRPO, along with the 7B merge revision model, to illustrate the model's step-by-step reasoning process on BIRD development set.

\subsection{Some successful merge revision cases}
\label{sec:some_success_merge_revision_case}

The examples include outputs from the top-1 group (where the SQL generation is incorrect) and the top-2 group (where the SQL generation is correct) after SC voting, as well as the final top-1 output produced by CSC, which demonstrates that CSC can effectively correct errors made by the SC method. The red part of the CSC output shows its correction process.

\onecolumn

\subsubsection{Successful Case 82}
\label{sec:some_success_case_82}

%  example 82 gold
\begin{tcolorbox}[colback=mainboxbg, colframe=mainboxborder, coltitle=black, fonttitle=\bfseries, title=The gold sql for question 82 ,bottom=5mm, boxrule=0.5mm, arc=2mm, outer arc=2mm]
question\_id: 82 \\
db\_id: california\_schools\\
question: What is the grade span offered in the school with the highest longitude?\\
evidence: the highest longitude refers to the school with the maximum absolute longitude value.\\
difficulty: simple\\
gold sql: SELECT GSoffered FROM schools ORDER BY ABS(longitude) DESC LIMIT 1\\
\end{tcolorbox}

%  example 82 sc top1 group output
\begin{tcolorbox}[colback=mainboxbg, colframe=mainboxborder, coltitle=black, fonttitle=\bfseries, title=The top 1 group results after SC voting for question 82 ,bottom=5mm, boxrule=0.5mm, arc=2mm, outer arc=2mm]
<think> \\
To answer this question, we need to find the school with the highest longitude value and retrieve its grade span offered. Here are the steps to derive the SQL query: \\

1. **Identify the relevant tables**: We need to access the `schools` table to get the longitude values and the grade span offered. The `Longitude` column is in the `schools` table, and the `GSoffered` column is also in the same table. \\

2. **Order the schools by longitude in descending order**: To find the school with the highest longitude, we need to order the schools by the `Longitude` column in descending order. This will put the school with the highest longitude at the top. \\

3. **Limit the result to the school with the highest longitude**: Since we are only interested in the school with the highest longitude, we can limit the result to just 1 row. \\

4. **Select the grade span offered**: Finally, we only need to select the `GSoffered` column from the `schools` table. \\

Putting it all together, the SQL query will select the `GSoffered` column from the `schools` table, order the results by `Longitude` in descending order, and limit the result to 1 row. \\
</think> \\
<answer> \\
SELECT GSoffered FROM schools ORDER BY Longitude DESC LIMIT 1 \\
</answer> \\

\end{tcolorbox}

%  example 82 sc top2 group output
\begin{tcolorbox}[colback=mainboxbg, colframe=mainboxborder, coltitle=black, fonttitle=\bfseries, title=The top 2 group results after SC voting for question 82 ,bottom=5mm, boxrule=0.5mm, arc=2mm, outer arc=2mm]

<think> \\
1. **Identify the relevant tables and columns**: The question asks for the grade span offered in the school with the highest longitude. The school information, including its longitude, is stored in the `schools` table. The grade span offered is stored in the `GSoffered` column of the same table. \\

2. **Determine the order and limit**: To find the school with the highest longitude, we need to order the `schools` table by the `Longitude` column in descending order and limit the result to the first row using the `LIMIT 1` clause. \\

3. **Construct the SQL query**: The SQL query will select the `GSoffered` column from the `schools` table, order the results by the `Longitude` column in descending order, and limit the output to the first row. \\
</think> \\
<answer> \\
SELECT GSoffered FROM schools ORDER BY ABS(Longitude) DESC LIMIT 1 \\
</answer> \\
\end{tcolorbox}

%  example 82 csc top1 group output
\begin{tcolorbox}[colback=mainboxbg, colframe=mainboxborder, coltitle=black, fonttitle=\bfseries, title=The top 1 group results after CSC voting for question 82 ,bottom=5mm, boxrule=0.5mm, arc=2mm, outer arc=2mm]
<think> \\
To answer the question "What is the grade span offered in the school with the highest longitude?", we need to follow these steps: \\

1. **Understand the Data**: The table `schools` contains information about schools, including their grade span (`GSoffered`) and longitude (`Longitude`). \\

2. **Identify the Requirement**: We need to find the school with the highest longitude value and then retrieve its grade span. \\

3. **Determine the SQL Operation**: \\
   - **ORDER BY**: To sort the schools by longitude in descending order. \\
   - **LIMIT**: To get only the top result, which corresponds to the school with the highest longitude. \\
   - **SELECT**: To specify that we want to retrieve the `GSoffered` column from the top result. \\

4. ** distinguish between Absolute Longitude**: The question asks for the "highest" longitude, which could mean either the maximum or minimum longitude value. In this context, it seems to be asking for the maximum absolute value of longitude, which could be either positive or negative. The first draft SQL query uses `ABS(Longitude)` to achieve this, sorting by the absolute value of longitude in descending order. However, if the intention is to find the school with the most eastern longitude (positive value), we should use `Longitude` directly. \\

5. **Construct the SQL Query**: \\
   - **SELECT**: We need to select the `GSoffered` column. \\
   - **FROM**: We need to specify the `schools` table. \\
   - **ORDER BY**: We sort by `Longitude DESC` to get the school with the highest longitude at the top. \\
   - **LIMIT**: We limit the results to 1 to get only the top result. \\

\textcolor{red}{The draft SQL query provided in the instruction already correctly follows these steps. It uses `ABS(Longitude)` to find the school with the highest absolute longitude value, which is the intended meaning of "highest longitude" in this context. }\\

Therefore, the correct SQL query is: \\
</think> \\
<answer> \\
SELECT GSoffered FROM schools ORDER BY \textcolor{red}{ABS(Longitude)} DESC LIMIT 1 \\
</answer> \\
\end{tcolorbox}

\subsubsection{Successful Case 232}
\label{sec:some_success_case_232}

%  example 232 gold
\begin{tcolorbox}[colback=mainboxbg, colframe=mainboxborder, coltitle=black, fonttitle=\bfseries, title=The gold sql for question 232 ,bottom=5mm, boxrule=0.5mm, arc=2mm, outer arc=2mm]
question\_id: 232 \\
db\_id: toxicology \\
question: Please list top three molecules that have single bonds between two atoms and are not carcinogenic in alphabetical order. \\
evidence: label = '-' means molecules are not carcinogenic; single type bond refers to bond\_type = '-'; list top three molecules refers to return molecule\_id and order by molecule\_id; \\
difficulty: moderate \\
gold sql: SELECT DISTINCT T2.molecule\_id FROM bond AS T1 INNER JOIN molecule AS T2 ON T1.molecule\_id = T2.molecule\_id WHERE T1.bond\_type = '-' AND T2.label = '-' ORDER BY T2.molecule\_id LIMIT 3 \\
\end{tcolorbox}

%  example 232 sc top1 group output
\begin{tcolorbox}[colback=mainboxbg, colframe=mainboxborder, coltitle=black, fonttitle=\bfseries, title=The top 1 group results after SC voting for question 232 ,bottom=5mm, boxrule=0.5mm, arc=2mm, outer arc=2mm]
<think>\\
1. **Identify the key requirements**: The question asks for the top three molecules that have single bonds between two atoms and are not carcinogenic. We need to find the molecules based on these conditions and order them alphabetically by `molecule\_id`.\\

2. **Determine the relevant tables**: \\
   - `molecule` table contains the ` molecule\_id` and `label` of molecules.\\
   - `bond` table contains the `bond\_type` of bonds and is linked to `molecule\_id`.\\

3. **Filter conditions**:\\
   - We need to filter for molecules where `label = '-'` (not carcinogenic).\\
   - We need to filter for bonds where `bond\_type = '-'` (single bond).\\

4. **Join the tables**:\\
   - Join `molecule` and `bond` tables on `molecule\_id`.\\

5. **Order and limit**:\\
   - Order the results by `molecule\_id` and limit the output to the top three.\\

6. **Formulate the SQL query**:\\
   - Use an INNER JOIN to combine `molecule` and `bond` tables.\\
   - Apply WHERE clauses to filter for not carcinogenic and single bonds.\\
   - Use ORDER BY to sort by `molecule\_id`.\\
   - Use LIMIT to restrict the output to 3 rows.\\
</think>\\
<answer>\\
SELECT T1.molecule\_id FROM molecule AS T1 INNER JOIN bond AS T2 ON T1.molecule\_id = T2.molecule\_id WHERE T1.label = '-' AND T2.bond\_type = '-' ORDER BY T1.molecule\_id LIMIT 3\\
</answer>\\
\end{tcolorbox}

%  example 232 sc top2 group output
\begin{tcolorbox}[colback=mainboxbg, colframe=mainboxborder, coltitle=black, fonttitle=\bfseries, title=The top 2 group results after SC voting for question 232 ,bottom=5mm, boxrule=0.5mm, arc=2mm, outer arc=2mm]
<think>\\
To answer this question, we need to find the top three molecules that have single bonds between two atoms and are not carcinogenic. Here's a step-by-step approach to derive the SQL query:\\

1. **Identify the relevant tables**: We need to access information from the `molecule`, `bond`, and `connected` tables. The `molecule` table contains the `label` information, the `bond` table contains the `bond\_type` information, and the `connected` table links atoms to bonds and molecules.\\

2. **Filter molecules not carcinogenic**: We will use the `WHERE` clause to filter molecules with `label = '-'` (not carcinogenic).\\

3. **Filter single type bonds**: We will also use a `WHERE` clause to filter bonds with `bond\_type = '-'` (single bonds).\\

4. **Group by molecule\_id**: Since we want to list the top three molecules, we will use the `GROUP BY` clause to group the results by `molecule\_id`.\\

5. **Order by molecule\_id**: Finally, we will use the `ORDER BY` clause to sort the results in alphabetical order.\\

Here's how we can construct the SQL query:\\

- Start with the `SELECT` statement to specify the required columns (`molecule\_id`).\\
- Use `FROM` and `INNER JOIN` to join the `molecule`, `bond`, and `connected` tables based on the `molecule\_id` and `bond\_id`.\\
- Add the `WHERE` clauses to filter molecules not carcinogenic and single type bonds.\\
- Add the `GROUP BY` clause to group the results by `molecule\_id`.\\
- Add the `ORDER BY` clause to sort the results in alphabetical order.\\
- Use the `LIMIT 3` clause to return only the top three molecules.\\
</think>\\
<answer>\\
SELECT T1.molecule\_id FROM molecule AS T1 INNER JOIN bond AS T2 ON T1.molecule\_id = T2.molecule\_id WHERE T1.label = '-' AND T2.bond\_type = '-' GROUP BY T1.molecule\_id ORDER BY T1.molecule\_id ASC LIMIT 3\\
</answer>\\
\end{tcolorbox}

%  example 232 csc top1 group output
\begin{tcolorbox}[colback=mainboxbg, colframe=mainboxborder, coltitle=black, fonttitle=\bfseries, title=The top 1 group results after CSC voting for question 232 ,bottom=5mm, boxrule=0.5mm, arc=2mm, outer arc=2mm]
<think>\\
The task requires us to find the top three molecules that have single bonds and are not carcinogenic, ordered alphabetically by molecule\_id. The correct SQL query should:\\
1. Join the `molecule` and `bond` tables based on the `molecule\_id` column.\\
2. Filter the joined data to include only those rows where the `label` is '-' (indicating the molecule is not carcinogenic) and the `bond\_type` is '-' (indicating a single bond).\\
3. Group the results by `molecule\_id` to exclude duplicates in the result set.\\
4. Order the results by `molecule\_id` in ascending order.\\
5. Limit the results to the top 3 rows.\\

\textcolor{red}{The draft SQL query 1 seems to have an issue; it's not correctly filtering out duplicates and isn't grouping the results, which would explain why it returns duplicate rows. The draft SQL query 2 avoids duplicates by grouping by `molecule\_id` but still isn't excluding duplicate rows, which could be due to a syntax error or oversight.\\}

\textcolor{red}{To correct the query, we need to ensure that the `GROUP BY` clause correctly groups the results, but since we're only interested in the `molecule\_id`, we should only select that column in the `SELECT` clause. We also need to ensure that the `LIMIT` clause correctly restricts the output to the top 3 rows.\\}
</think>\\
<answer>\\
SELECT T1.molecule\_id FROM molecule AS T1 INNER JOIN bond AS T2 ON T1.molecule\_id = T2.molecule\_id WHERE T1.label = '-' AND T2.bond\_type = '-' \textcolor{red}{GROUP BY T1.molecule\_id} ORDER BY T1.molecule\_id ASC LIMIT 3\\
</answer>\\
\end{tcolorbox}  

\subsection{Some failed merge revision cases}
\label{sec:some_failed_merge_revision_case}

\textcolor{red}{We also found some failure cases where the merged revision model did not output the final result correctly, which shows that the merge revision model also has certain limitations and does not work in some situations. In the future, we will further analyze these failure cases to improve the merge revision model.} 

\subsubsection{Failed Case 15}
\label{sec:some_failed_case_15}

%  example 15 gold
\begin{tcolorbox}[colback=mainboxbg, colframe=mainboxborder, coltitle=black, fonttitle=\bfseries, title=The gold sql for question 15 ,bottom=5mm, boxrule=0.5mm, arc=2mm, outer arc=2mm]
question\_id: 15\\
db\_id: california\_schools\\
question: Which active district has the highest average score in Reading?\\
evidence:
difficulty: simple\\
gold sql: SELECT T1.District FROM schools AS T1 INNER JOIN satscores AS T2 ON T1.CDSCode = T2.cds WHERE T1.StatusType = 'Active' ORDER BY T2.AvgScrRead DESC LIMIT 1\\
\end{tcolorbox}

%  example 15 sc top1 group output
\begin{tcolorbox}[colback=mainboxbg, colframe=mainboxborder, coltitle=black, fonttitle=\bfseries, title=The top 1 group results after SC voting for question 15 ,bottom=5mm, boxrule=0.5mm, arc=2mm, outer arc=2mm]

<think>\\
To answer the question, we need to follow these steps:\\

1. **Identify the relevant tables**: The question involves finding districts with high SAT scores and determining which one is active. Therefore, we need to look at the `satscores` table for SAT scores and the `schools` table for district information.\\

2. **Join the tables**: Since the `satscores` table doesn't directly contain district details, we need to join it with the `schools` table using the `CDSCode` column, which is the primary key in `schools` and acts as a foreign key in `satscores`.\\

3. **Filter for active districts**: We need to filter the results to include only active districts, which can be identified by the `StatusType` column in the `schools` table.\\

4. **Calculate the average reading score**: We need to calculate the average reading score for each active district.\\

5. **Order by average score**: We need to order the results by the average reading score in descending order to find the district with the highest average score.\\

6. **Limit the results**: Since we are only interested in the district with the highest average score, we can limit the results to 1.\\

7. **Select the district name**: Finally, we only need to select the district name, which is the `District` column in the `schools` table.\\

Given these steps, the SQL query can be constructed as follows:\\
</think>\\
<answer>\\
SELECT T2.District FROM satscores AS T1 INNER JOIN schools AS T2 ON T1.cds = T2.CDSCode WHERE T2.StatusType = 'Active' GROUP BY T2.District ORDER BY AVG(T1.AvgScrRead) DESC LIMIT 1\\
</answer>\\
\end{tcolorbox}

%  example 15 sc top2 group output
\begin{tcolorbox}[colback=mainboxbg, colframe=mainboxborder, coltitle=black, fonttitle=\bfseries, title=The top 2 group results after SC voting for question 15 ,bottom=5mm, boxrule=0.5mm, arc=2mm, outer arc=2mm]
<think>\\
To answer this question, we need to follow these steps:\\

1. **Identify the relevant tables**: The question involves finding information about districts and their satscores. Therefore, the relevant tables are `schools` and `satscores`.\\

2. **Understand the relationships**: The `schools` table has a foreign key `CDSCode` that references the `CDSCode` in the `satscores` table. This means we can join these two tables on the `CDSCode` column.\\

3. **Filter active districts**: The question specifies that we should only consider active districts. Therefore, we need to add a filter in the `WHERE` clause to include only `StatusType = 'Active'` districts.\\

4. **Sort by average score in Reading**: To find the district with the highest average score in Reading, we need to sort the results in descending order based on the `AvgScrRead` column from the `satscores` table.\\

5. **Limit the output**: Since we are interested in the district with the highest average score, we can limit the output to the first row.\\

6. **Select the required columns**: The question asks for the district name, so we need to select the `dname` column from the `satscores` table.\\

Based on these steps, the SQL query can be written as follows:\\
</think>\\
<answer>\\
SELECT T2.dname FROM schools AS T1 INNER JOIN satscores AS T2 ON T1.CDSCode = T2.cds WHERE T1.StatusType = 'Active' ORDER BY T2.AvgScrRead DESC LIMIT 1\\
</answer>\\
\end{tcolorbox}

%  example 15 csc top1 group output
\begin{tcolorbox}[colback=mainboxbg, colframe=mainboxborder, coltitle=black, fonttitle=\bfseries, title=The top 1 group results after CSC voting for question 15 ,bottom=5mm, boxrule=0.5mm, arc=2mm, outer arc=2mm]
<think>\\
To answer the question "Which active district has the highest average score in Reading?", we need to follow these steps:\\

1. **Identify the relevant tables and columns**: \\
   - `satscores` table: contains the average reading scores.\\
   - `schools` table: contains the districts and their status.\\

2. **Join the tables**: \\
   - We need to join the `satscores` table with the `schools` table on the `cds` column to link each district with its average reading score.\\

3. **Filter for active districts**: \\
   - We need to filter the results to include only active districts. This can be done using a WHERE clause.\\

4. **Group by district**: \\
   - We need to group the results by district to calculate the average score for each district.\\

5. **Order by average score in descending order**: \\
   - We need to sort the results by the average score in descending order to get the district with the highest score at the top.\\

6. **Limit the results**: \\
   - Since we only want the district with the highest average score, we can limit the results to 1.\\

\textcolor{red}{According to the draft SQL provided, the steps seem to be followed correctly, but the table aliases in the query are not consistent. We will correct them to ensure the query works.}\\

The correct SQL query should look like this:\\
</think>\\
<answer>\\
SELECT T2.District FROM satscores AS T1 INNER JOIN schools AS T2 ON T1.cds = T2.CDSCode WHERE T2.StatusType = 'Active' GROUP BY T2.District ORDER BY AVG(T1.AvgScrRead) DESC LIMIT 1\\
</answer>\\
\end{tcolorbox}

\subsubsection{Failed Case 590}
\label{sec:some_failed_case_590}

%  example 590 gold
\begin{tcolorbox}[colback=mainboxbg, colframe=mainboxborder, coltitle=black, fonttitle=\bfseries, title=The gold sql for question 590 ,bottom=5mm, boxrule=0.5mm, arc=2mm, outer arc=2mm]
question\_id: 590\\
db\_id: codebase\_community\\
question: Which user ID has the lowest view?\\
evidence: lowest views refers to Min(Views)\\
difficulty: simple\\
gold sql: SELECT Id FROM users WHERE Views = ( SELECT MIN(Views) FROM users )\\
\end{tcolorbox}

%  example 590 sc top1 group output
\begin{tcolorbox}[colback=mainboxbg, colframe=mainboxborder, coltitle=black, fonttitle=\bfseries, title=The top 1 group results after SC voting for question 590 ,bottom=5mm, boxrule=0.5mm, arc=2mm, outer arc=2mm]
<think>\\
To answer the question "Which user ID has the lowest view?", we need to follow these steps:\\

1. **Identify the relevant table**: The question is about users, their views, and their IDs. The most relevant table for this information is the `users` table.\\

2. **Determine the relevant columns**: The columns of interest are `Views` and `Id` (user ID).\\

3. **Order the results by views**: To find the user with the lowest views, we need to order the results in ascending order based on the `Views` column.\\

4. **Limit the results**: Since we are only interested in the user with the lowest views, we can limit the results to the first row of the ordered list.\\

5. **Select the correct column**: The question asks for the user ID, so we need to select the `Id` column.\\

Putting these steps together, we can formulate the SQL query.\\
</think>\\
<answer>\\
SELECT Id FROM users ORDER BY Views ASC LIMIT 1\\
</answer>\\
\end{tcolorbox}

%  example 590 csc top1 group output
\begin{tcolorbox}[colback=mainboxbg, colframe=mainboxborder, coltitle=black, fonttitle=\bfseries, title=The top 1 group results after CSC voting for question 590 ,bottom=5mm, boxrule=0.5mm, arc=2mm, outer arc=2mm]

<think>
To answer the question "Which user ID has the lowest view?", we need to find the user with the minimum number of views. Here are the steps to derive the SQL query:\\

1. Identify the relevant table: The `users` table contains information about users, including their views.\\
2. Determine the column of interest: The `Views` column in the `users` table contains the number of views each user has.\\
3. Find the minimum value of views: To find the user with the lowest view, we need to identify the user with the minimum value in the `Views` column.\\
4. Select the user ID: Once we have the minimum value, we select the user ID from the `users` table that corresponds to this minimum value.\\

The SQL query will use the `MIN` function to find the minimum number of views and then use a subquery to find the user ID with this minimum value.\\
</think>\\
<answer>\\
SELECT Id FROM users WHERE Views = ( SELECT MIN(Views) FROM users );\\
</answer>\\
 \end{tcolorbox}

 %  example 590 csc top1 group output
\begin{tcolorbox}[colback=mainboxbg, colframe=mainboxborder, coltitle=black, fonttitle=\bfseries, title=The top 1 group results after CSC voting for question 590 ,bottom=5mm, boxrule=0.5mm, arc=2mm, outer arc=2mm]
<think>\\
Let's analyze the question and the database schema. The question asks for the user ID with the lowest view count. The database schema has a table named `users` with two columns: `Id` and `Views`.\\

1. **Identify the required operation**: The task requires finding the user ID with the minimum view count.\\
2. **Determine the appropriate SQL function**: The SQL function `MIN` can be used to find the minimum value in the `Views` column.\\
3. **Formulate the SQL query**:\\
   - We need to select the `Id` from the `users` table.\\
   - We use `ORDER BY Views ASC` to sort the rows in ascending order based on the view count.\\
   - We use `LIMIT 1` to get only the top row, which corresponds to the user with the lowest view count.\\
\textcolor{red}{4. **Validate the query**:\\
   - The first draft SQL provided is: `SELECT Id FROM users ORDER BY Views ASC LIMIT 1`\\
   - The execution result confirms that this query correctly returns the user ID with the lowest view count.\\}

Based on these steps, the final SQL query is: `SELECT Id FROM users ORDER BY Views ASC LIMIT 1`\\
</think>\\
<answer>\\
SELECT Id FROM users ORDER BY Views ASC LIMIT 1\\
</answer>
 \end{tcolorbox}

% \newpage
\section{Prompt Templates}
\label{sec:appendix_prompt}

\onecolumn
\subsection{Prompt for SQL Generation}
\label{sec:prompt_candidate_sql_generate}

\begin{tcolorbox}[colback=mainboxbg, colframe=mainboxborder, coltitle=black, fonttitle=\bfseries, title=SQL Generation Prompt,bottom=60mm, boxrule=0.5mm, arc=2mm, outer arc=2mm]
    You first thinks about the reasoning process in the mind and then provides the user with the answer.\\

    Task Overview:\\
    You are a data science expert. Below, you are provided with a database schema and a natural language question. Your task is to understand the schema and generate a valid SQL query to answer the question.\\
    
    Database Engine:\\
    SQLite\\
    
    Database Schema:\\
    \{DATABASE SCHEMA\}\\
    \\
    This schema describes the database's structure, including tables, columns, primary keys, foreign keys, and any relevant relationships or constraints.\\

    Question:\\
    \{EVIDENCE\}\\
    \{QUESTION\}\\
    
    Instructions:\\
    \mbox{-} Make sure you only output the information that is asked in the question. If the question asks for a specific column, make sure to only include that column in the SELECT clause, nothing more.\\
    \mbox{-} The generated query should return all of the information asked in the question without any missing or extra information.\\
    \mbox{-} Before generating the final SQL query, please think through the steps of how to write the query.\\
    
    Output Format:\\
    Show your work in <think> </think> tags. And return the final SQLite SQL query that starts with keyword `SELECT` in <answer> </answer> tags, for example <answer>SELECT AVG(rating\_score) FROM movies</answer>.  \\
    
    Let me solve this step by step.\\
\end{tcolorbox}

\subsection{Prompt for SQL Merge Revision}
\label{sec:prompt_sql_merge_revision}

\begin{tcolorbox}[colback=mainboxbg, colframe=mainboxborder, coltitle=black, fonttitle=\bfseries, title=SQL Merge Revision Prompt,bottom=1mm, boxrule=0.5mm, arc=2mm, outer arc=2mm]

    You first thinks about the reasoning process in the mind and then provides the user with the answer.\\

    Task Overview:\\
    You are a data science expert. Below, you are provided with a database schema, a natural language question, some draft SQL and its corresponding execution result. Your task is to understand the schema and generate a valid SQL query to answer the question.\\

    Database Engine:\\
    SQLite\\

    Database Schema:\\
    \{DATABASE SCHEMA\}\\
    \\
    This schema describes the database's structure, including tables, columns, primary keys, foreign keys, and any relevant relationships or constraints.\\

    Question:\\
    \{EVIDENCE\}\\
    \{QUESTION\}\\

    Here are some corresponding draft SQL and execute result:\\
    1. \{PREDICT\_SQL1\} \\
    Execution result\\
    \{EXECUTE\_RESULT1\}\\

    2. \{PREDICT\_SQL2\} \\
    Execution result\\
    \{EXECUTE\_RESULT2\}\\

    Instructions:\\
    \mbox{-} Make sure you only output the information that is asked in the question. If the question asks for a specific column, make sure to only include that column in the SELECT clause, nothing more.\\
    \mbox{-} The generated query should return all of the information asked in the question without any missing or extra information.\\
    \mbox{-} Before generating the final SQL query, please think through the steps of how to write the query.\\
    
    Output Format:\\
    Show your work in <think> </think> tags. And return the final SQLite SQL query that starts with keyword `SELECT` in <answer> </answer> tags, for example <answer>SELECT AVG(rating\_score) FROM movies</answer>. \\ 
    
    Let me solve this step by step.\\
\end{tcolorbox}

\end{document}